\documentclass[letterpaper, 10 pt, conference]{ieeeconf}  

\IEEEoverridecommandlockouts                              

\usepackage{amsmath}
\usepackage{amssymb}             
\usepackage{amsfonts}            
\usepackage{mathrsfs}            
\usepackage{graphicx}
\usepackage{float}
\usepackage{subfig}
\usepackage{booktabs}

\title{\LARGE \bf
Stylized Table Tennis Robots Skill Learning with Incomplete Human Demonstrations
}

\author{Xiang Zhu*$^{1}$\and 
Zixuan Chen*$^{2,3}$
\and Jianyu Chen$^{1,3}$
\thanks{*denotes equal contribution, order is determined by roll of a dice}
\thanks{$^{1}$Tsinghua University, $^{2}$Fudan University, $^{3}$Shanghai Qi Zhi Institute}%
}

\begin{document}

\maketitle
\thispagestyle{empty}
\pagestyle{empty}

\begin{abstract}
In recent years, Reinforcement Learning (RL) is becoming a popular technique for training controllers for robots. However, for complex dynamic robot control tasks, RL-based method often produces controllers with unrealistic styles. In contrast, humans can learn well-stylized skills under supervisions. For example, people learn table tennis skills by imitating the motions of coaches. Such reference motions are often incomplete, e.g. without the presence of an actual ball. Inspired by this, we propose an RL-based algorithm to train a robot that can learn the playing style from such incomplete human demonstrations. We collect data through the teaching-and-dragging method. We also propose data augmentation techniques to enable our robot to adapt to balls of different velocities. We finally evaluate our policy in different simulators with varying dynamics.
\end{abstract}

\section{Introduction}


Reinforcement Learning (RL) has emerged as a powerful technique for training intelligent agents in various domains, including sport games. Table tennis, often referred to as ping pong, is a sport demanding rapid reflexes and dynamic control. This poses a significant challenge for RL algorithms, which can exhibit unrealistic behaviors, unlike human players.

Traditionally, humans acquire the skill of table tennis through fragmentary and incomplete demonstrations. For instance, in the context of instruction, a coach might manipulate a trainee's arm to exemplify the appropriate striking mechanism, notwithstanding the absence of a ball during this demonstration. Yet, humans are still able to extract the essence of such motions to proficiently play the game.  These demonstrations provide valuable insights into the desired style of playing, that is the swing motion used to hit the ball. Our research aims to use these partial demonstrations to develop an RL-based system for a robotic arm to play table tennis.

While previous research has utilized RL techniques for playing ping-pong, none of them have specifically focused on learning the style of play from incomplete and partial demonstrations. Most existing approaches rely on direct RL training, where the agent learns to optimize its actions solely through trial and error solely based on the reward functions designed manually by humans. While \cite{gao2020robotic} made strides in addressing style, their defined style is not readily extensible. Our proposed approach differs by integrating knowledge from human demonstrations, allowing the agent to adopt specific playing styles.

In this paper we create a stylized table tennis agent that learns from partial demonstrations, even without ball trajectory data. Using generative adversarial learning techniques and insights from the teaching-and-dragging method, our goal is to develop an agent that demonstrates unique playing styles and efficiently returns the ball. We gather demonstrations by guiding the robotic arms through various swing motions and subsequently use these demonstrations to train an RL policy. We illustrate that our policy has the potential to be transferred to a real robot through a sim2sim task.

\section{Related Work}

\noindent\textbf{Reinforcement Learning for Table Tennis}\quad In recent years,  the application of reinforcement learning (RL) to robotic arm control in table tennis has garnered significant attention. Comprehensive overviews are provided by \cite{mahjourian2019hierarchical} and \cite{dambrosio2023robotic}, highlighting the advancements in high-speed table tennis RL systems. Numerous studies, including \cite{monte2018zhu}, \cite{optstrike2015huang}, \cite{balance2011sun}, \cite{gao2022optimal}, and \cite{gao2022model}, have focused on designing RL algorithms to improve ball-return accuracy and speed, specifically by predicting ball state at time $t$ to determine optimal paddle positions for striking by learning methods. \cite{gao2020robotic} focused on using model-free RL algorithms to develop control policies for robotic arm joint velocities in table tennis.  \cite{tebbe2021sample}  employed a one-step environment with well-defined action space and controller, aiming for high sample-efficiency. \cite{yang2021ball}  introduced a robotic simulation environment, rooted in mathematical modeling, to predict ball spin velocities using RL. In a novel approach, \cite{buchler2022learning}  utilized pneumatic artificial muscles (PAMs) and end-to-end RL to formulate hitting strategies. They also introduced a hybrid simulation-real training (HYSR) method to optimize safety and sample efficiency. \cite{ding2022goalseye} aimed to return balls to specific landing positions, while \cite{abeyruwan2023sim2real} implemented an iterative sim2real system, enabling trained agents to compete against humans.

\noindent\textbf{Reinforcement Learning from Demonstrations}\quad Advancements in robotics have led to numerous applications demanding intelligent systems capable of decision-making and practical physical movements. Nonetheless, sub-optimal parameter configurations or algorithmic constraints might prevent a learning agent from achieving the desired behavior \cite{pathak2019self}, thus hindering deployment real-world deployment. While task learning can technically be addressed as an optimization through meticulous reward engineering, integrating expert-driven prior knowledge is often deemed more efficient than starting anew \cite{billard2008survey}. One recognized approach to learning from demonstrations is inverse reinforcement learning \cite{schaal1996learning}, wherein an agent's goals or preferences are inferred from its actions rather than from direct reward feedback. In the era of deep learning, by combining the power of generative neural networks and imitation learning \cite{goodfellow2014generative}, generative adversarial imitation learning (GAIL)\cite{ho2016generative} is proposed to directly extract policy from demonstration data. In robotics, demonstrations is a very common and useful form of reference motions. They can be learned through tracking objectives that minimize pose error between simulated characters and target motions \cite{liu2010sampling}\cite{liu2016guided}\cite{peng2018deepmimic}. However, GAIL struggles in environments with substantial deviations from demonstrations, given the challenge of balancing imitation with task accomplishment \cite{fu2017learning}. To this end, \cite{peng2021amp} introduced the adversarial motion prior (AMP) that utilizes a discriminator to capture the distribution of input mocap without exact imitation. Subsequent works \cite{escontrela2022adversarial} and \cite{wang2023amp} proved that AMP's efficacy on real robots and that it can be powerful for solving locomotion tasks. Further enhancements were made by \cite{li2023learning}, integrating partial demonstrations and augmenting training stability.

We have noticed that prior research has predominantly focused on the ball-return task of table tennis, overlooking the significance of playing style and the specific actions involved. Existing methods, typically based on imitation learning or inverse RL, necessitate demonstrations with synchronized actions, speeds, and the ball's presence. In contrast, our study aims to infer playing styles from demonstrations captured by the arm's slow dragging movements in the absence of a ball, a context that may not reflect genuine gameplay dynamics.

\section{Method}
\begin{figure*}
    \centering
    \includegraphics[width=0.95\textwidth]{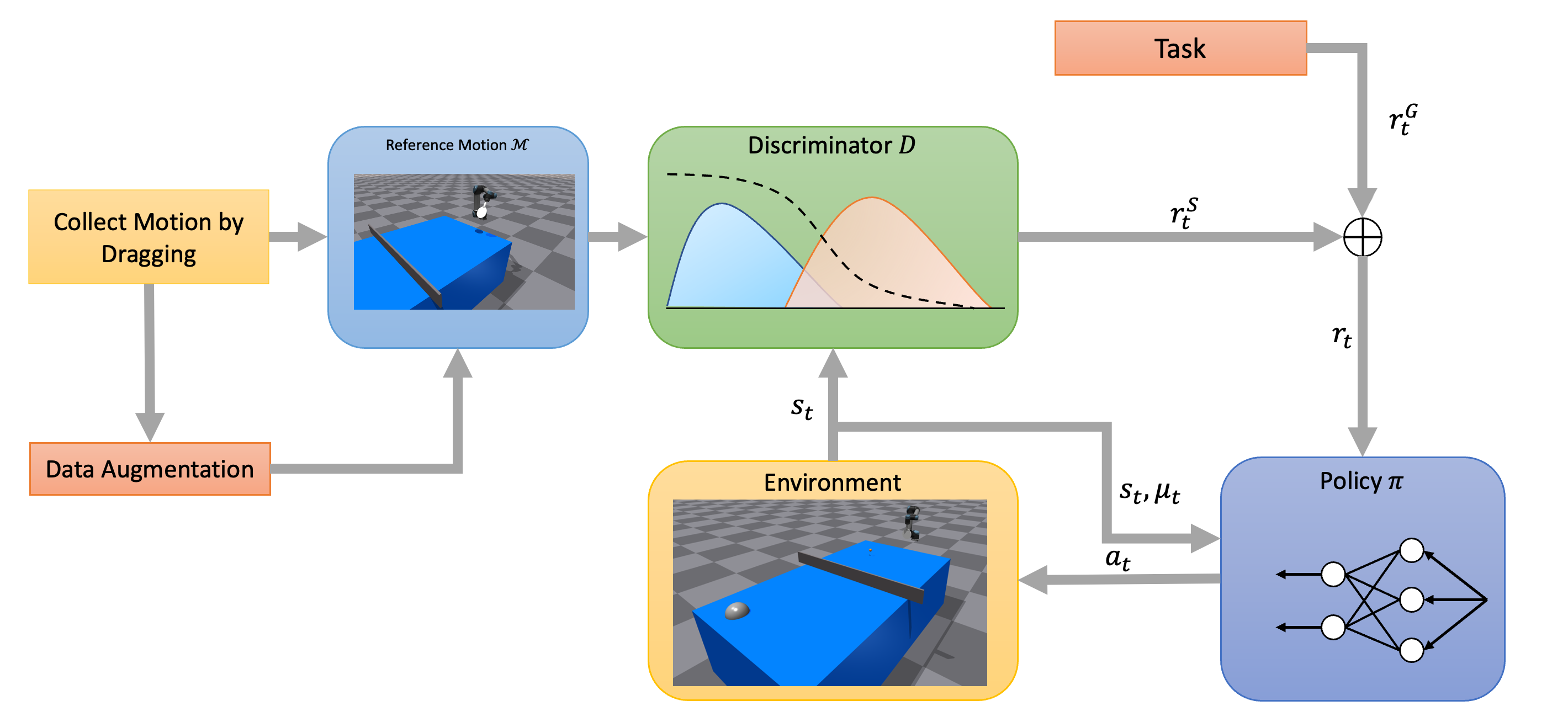}
    \caption{Overview of our method. We first collect reference motion data through manually guiding a robotic arm in a real-world environment. This data undergoes augmentation, producing our training reference motion \( \mathcal{M} \). Subsequently, this augmented reference motion, together with a predefined task, is utilized to concurrently train the discriminator and our policy.}
    \label{fig:shcematic}
\end{figure*}

Our objective is to derive robotic arm motion styles from human-guided dragging-and-teaching demonstrations to guide subsequent RL agent behaviors. We initially gather seven motion clips by manually dragging a UR5 robot. This data is then augmented by accelerating the reference motions at 2x, 3x, and 5x rates. Following this, we utilize the augmented motion data to train the discriminator in conjunction with policy and value networks based on the predefined task reward. A schematic representation of our methodology is presented in Figure \ref{fig:shcematic}.
\label{sec:method}

\subsection{Collecting Demonstration by Dragging}
We use Forward Dynamics Compliance Control (FDCC) \cite{scherzinger2017forward} to achieve the dragging-and-teaching procedure. FDCC combines the principles of admittance, impedance, and force control to achieve compliance control in Cartesian space. The core concept revolves around utilizing dynamic simulations to directly manipulate robotic manipulators through virtual and measured forces applied to their end effectors. By employing forward dynamics as a well-suited solver, the translation of effort from task space to motion commands in joint space is efficiently achieved. Here we define a Cartesian motion control target:
\begin{equation}
    F_{\text{t}} = K_P(x_d -x) - K_D\dot x
\end{equation}
where $F_t$ is the Cartesian task force, $x_d$ is the desired pose and $x$ is the current end effector pose. We then employ forward dynamics to determine the joint control targets.
\begin{equation}
     \ddot q= \text{ForwardDynamics}(\text{model},q,\dot q, F_{\text{ext}} - F_{\text{t}})
\end{equation}
where $q$ is the joint position, $F_{\text{ext}}$ is the external force. Then we can use integration to get $q$ and $\dot q$.

To collect demonstrations via dragging, we simply set the parameter \(K_P\) to 0, while the damping can be controlled by modifying the parameter \(K_D\). Following this, we guide the robotic arm to execute a specific swinging motion, capturing the robot's state \(s_{\text{arm}}\), including the joint positions \(q\) and velocities \(\dot q\). This accumulated data functions as reference motion for subsequent training. It is worth noting that, given the robot's intrinsic damping, the control strategy we adopted, and the absence of a ball during the dragging phase, a direct replay of the data from the dragging-and-teaching procedure would not result in a successful ping-pong ball return. The data collected comprise three primary components: forward, rightward, and leftward movements.
\subsection{Learning Styles from Partial Demonstrations}
In the context of the underlying Markov Decision Process (MDP), the state can be seen as two components: the robotic arm state and the ball state, represented as \(s = ( s_{arm}, s_{ball} )\). Utilizing the generative adversarial learning framework, the reward function is inferred for stylistic analysis from demonstrated transitions \( (s_{arm}, s'_{arm}) \) observed in a dragging-and-teaching demonstration. Within this framework, the policy functions as a generative model and stands in opposition to a discriminator. The role of the discriminator is to differentiate between transitions drawn from the reference demonstration distribution \(d^\mathcal{M}\) and those generated by the policy \(d^\mathcal{\pi}\).

In our approach, we adopt the AMP \cite{peng2021amp} framework, which employs the least-squares GAN (LSGAN) \cite{mao2017least} loss, to train the discriminator \(D(s_{arm}, s'_{arm})\). Distinct from the traditional GAN that commonly leverages a sigmoid cross-entropy loss function, the use of the LSGAN loss has been proved to enhance training stability and produce superior results in image synthesis tasks. For clarity in subsequent discussions, we omit the subscript from the discriminator input.
\begin{align}
\underset{D}{\arg \min } \; \mathbb{E}_{d^{\mathcal{M}}\left(\mathrm{s}, \mathrm{s}^{\prime}\right)}\left[\left(D\left(\mathrm{~s}, \mathrm{~s}^{\prime}\right)-1\right)^{2}\right] + \\ \mathbb{E}_{d^{\pi}\left(\mathrm{s}, \mathrm{s}^{\prime}\right)}\left[\left(D\left(\mathrm{~s}, \mathrm{~s}^{\prime}\right)+1\right)^{2}\right] .
\end{align}
In our method the discriminator is trained using the aforementioned loss, with the objective of predicting a score of $1$ for samples sourced from the demonstrations and a score of $-1$ for those generated by the policy. A commonly encountered challenge during GAN training is mode collapse, whereby the outputs produced by the generator demonstrate limited diversity, capturing only a constrained spectrum of possibilities. To mitigate this issue, we incorporate a gradient penalty \cite{gulrajani2017improved}. Additionally, due to the robotic arm's inherent limited degrees of freedom, the slow speed of the dragging-and-teaching process and its confined range of motion, the difference between two successive states can be quite small, which can also negatively affect the training of discriminator, increasing the probability of the discriminator to get trapped in local minima. We therefore take a state sequence of length $L$ as the input for the discriminator, denoted as $D(s_{t-L+1}, ..., s_{t})$.
\begin{equation}
\begin{aligned}
\underset{D}{\arg \min } \; & \mathbb{E}_{d^{\mathcal{M}}\left(s_{t-L+1}, ..., s_{t}\right)}
\left[\left(D\left(s_{t-L+1}, ..., s_{t}\right)-1\right)^{2}\right]+ \\
&\mathbb{E}_{d^{\pi}\left(s_{t-L+1}, ..., s_{t}\right)}\left[\left(D\left(s_{t-L+1}, ..., s_{t}\right)+1\right)^{2}\right] + \\
& \omega^{gp} \mathbb{E}_{d^{\mathcal{M}}\left(s_{t-L+1}, ..., s_{t}\right)}\left[ \lVert \nabla D_{\phi}(\phi)\rVert\right].
\label{eq:loss}
\end{aligned}
\end{equation}
The loss function for the discriminator is defined by Equation \ref{eq:loss}. Weight decay is also applied to stabilize the training procedure.
\subsection{Data Augmentation}
The reference motion obtained from drag-and-teach methods tends to be slow, making it ineffective against high-speed balls. Such a speed discrepancy can compromise the efficiency of imitation learning.  To ensure the system effectively learns the style from a slower reference motion while still performing optimally in high-speed ball interactions, we augment the original reference motion data. Let the original motion reference be represented as a sequence of joint poses and velocities, denoted by \( \{ \hat{q}_t, \hat{v}_t \} \), whose length is \( n \). To increase the motion speed by a factor of \( k \), we sample \( \frac{n}{k} \) joint poses at equidistant intervals to form a new motion reference, where the corresponding velocity is also scaled by \( k \). This yields a new reference motion, \( \{ \hat{q}_t^\prime, \hat{v}_t^\prime \} \). Both the augmented and the original motion references are later included in our demonstration dataset, collectively forming our augmented demonstration set.

\subsection{Reward Functions}
To facilitate the agent's proficiency in directing the table tennis ball to a specified location on the opposite of the table, we have formulated a reward system comprising three primary elements: ball hitting, smoothing \& regularization, and penalization for illegal penalties.

\noindent\textbf{Ball Hitting}\quad This reward encourages the agent to strike the ball and redirect it to the designated goal position.
\begin{equation}
r_{\text{hit}} =
\begin{cases}
\alpha_{\text{bat}}\exp\left( \lVert x_{\text{eff}_y} - x_{\text{ball}_y} \lVert / \sigma_{\text{bat}}\right),  & \text{\textbf{before} hitting,} \\
\alpha_{\text{goal}}\exp\left( \lVert g - x_{\text{ball}} \lVert / \sigma_{\text{goal}}\right), & \text{\textbf{after} hitting,}
\end{cases}
\end{equation}
where \(x_{\text{eff}_y}\) represents the \(y\) position of the arm's end effector, \(x_{\text{ball}_y}\) denotes the \(y\) position of the ping-pong ball, and \(g\) represents the goal, which is a predetermined ball landing location.

\noindent\textbf{Smoothing \& Regularization}\quad These rewards aim to smooth and regularize the agent's action.
\begin{itemize}
    \item Acceleration penalty, where $q$ is the robot joint position.
    \begin{equation}
        r_{\text{acc}} = \alpha_{\text{acc}} \exp{\left(\lVert \ddot q \rVert \right)}.
    \end{equation}

    \item Dof regulaztion, where $\hat{q}$ is the default joint positions, and $\sigma_{\text{dof}}$ is a predefined hyperparameter.
    \begin{equation}
        r_{\text{dof}} = \alpha_{\text{dof}} \exp{\left(\lVert q - \hat{q} \rVert \right / \sigma_{\text{dof}})}
    \end{equation}

    \item Action rate, where $a$ is the current action generated by the policy.
    \begin{equation}
        r_{\text{ar}} = \alpha_{\text{ar}} \exp{\left(\lVert a_t - a_{t-1} \rVert \right)}
    \end{equation}

    \item Action penalty, where $\mid a \mid$ is the absolute value of the current action.
    \begin{equation}
        r_{\text{ap}} = \alpha_{\text{ap}} \max{\mid a\mid}
    \end{equation}
\end{itemize}

\noindent\textbf{Illegal Penalties}\quad We also penalize illegal behaviors, such as: robot self collision, ball illegal bounce and etc.

The task reward $r_{t}^{G}$ is the summation of the above reward functions. The final reward consists of the task reward and the style reward. 
\begin{equation}
    r_t = \omega_{\text{task}} r_{t}^{G} + \omega_{\text{style}} r_{t}^{S},
\end{equation}
in which $ r_{t}^{S} = -\log \left(D\left(s_{t-L+1}, ..., s_{t}\right)\right)$.

\subsection{Domain Transfer}
To enhance the training process and strengthen our agent's adaptability to varied dynamics, we employ an asymmetric actor-critic network \cite{pinto2017asymmetric} with distinct policy and value inputs. The policy's input encompasses the information required and obtainable during the actual robot deployment. However, the value function incorporates additional privileged information, including the specific position and velocity of the agent, the number of bounces between the ball and the table, physical parameters of the environment, noise levels, damping values for robot velocity control, and delays in action and observation. These pieces of information are difficult to acquire during real robot deployment, hence we include them as additional inputs for the value function.

To achieve success during domain transfer, we incorporate extensive randomization of the environment’s dynamics parameters. We primarily consider four kinds of errors that could cause issues when transferring the policy to a real robot, including: discrepancies between the robot and environment models, errors in motor parameters, sensor noise, and communication delays. To enable the policy to account for discrepancies between the robot and environment models, we randomize the table, bat and the ball friction and restitution, gravity. We randomize the damping and stiffness parameter for the low-level position controller. We also heavily randomize the sensor noise which will input into the policy and also consider the communication and computation delays between the policy and input, as well as between the policy and action.

\begin{figure*}[ht]
    \centering
    \subfloat[ppo-step1]{\includegraphics[width=0.14\textwidth]{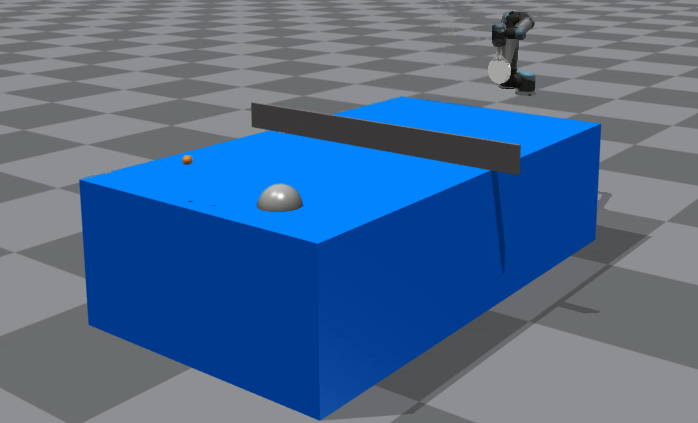}} 
    \subfloat[ppo-step2]{\includegraphics[width=0.14\textwidth]{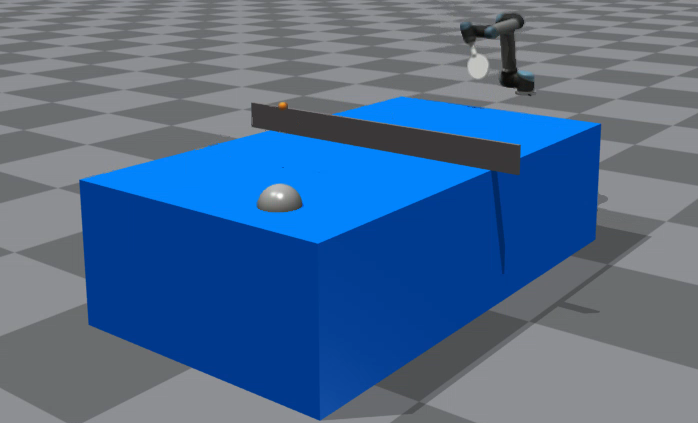}} 
    \subfloat[ppo-step3]{\includegraphics[width=0.14\textwidth]{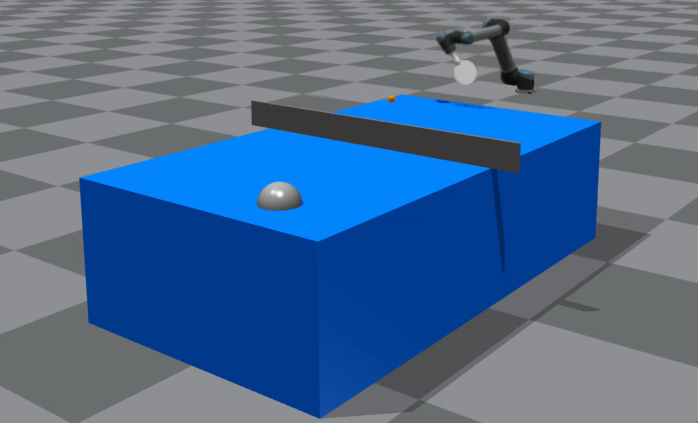}} 
    \subfloat[ppo-step4]{\includegraphics[width=0.14\textwidth]{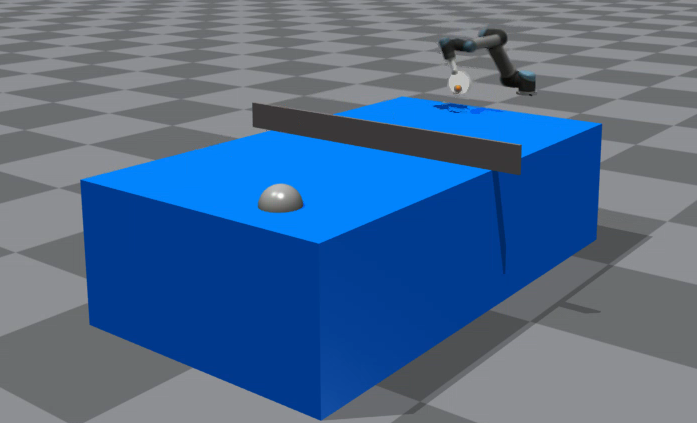}}
    \subfloat[ppo-step5]{\includegraphics[width=0.14\textwidth]{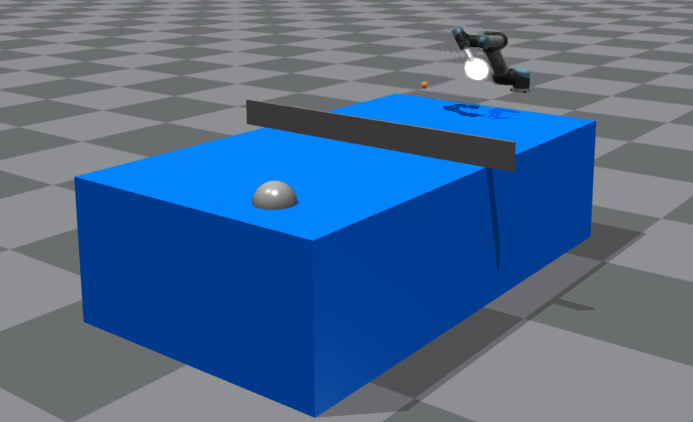}}
    \subfloat[ppo-step6]{\includegraphics[width=0.14\textwidth]{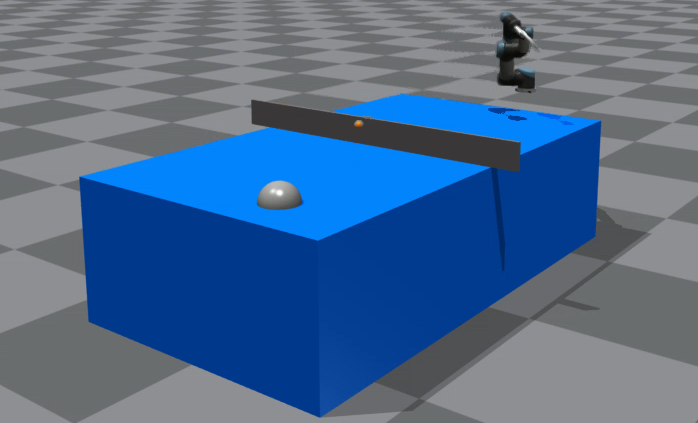}}
    \subfloat[ppo-step7]{\includegraphics[width=0.14\textwidth]{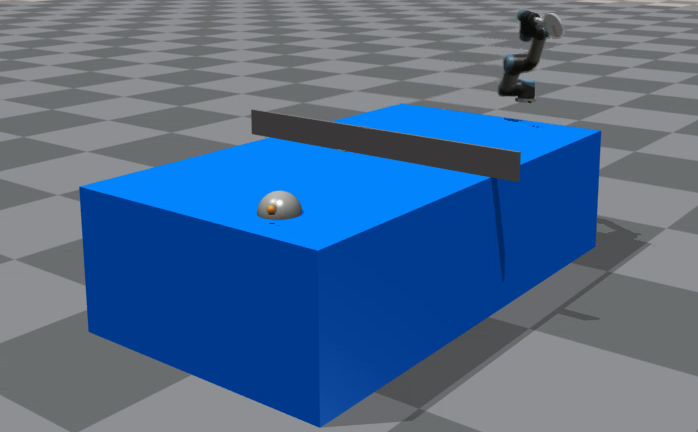}} \\
    \subfloat[amp-step1]{\includegraphics[width=0.14\textwidth]{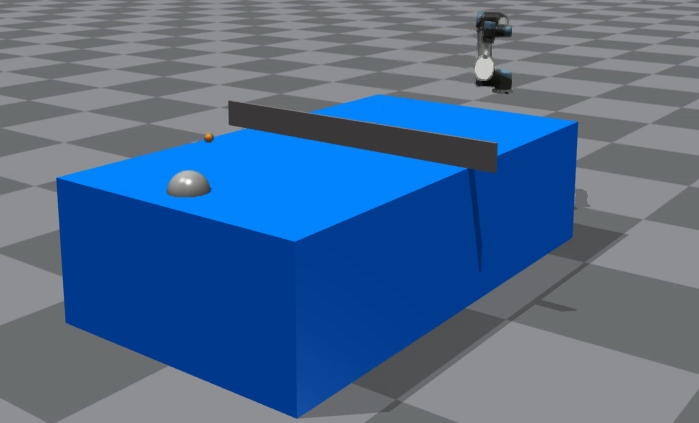}}
    \subfloat[amp-step2]{\includegraphics[width=0.14\textwidth]{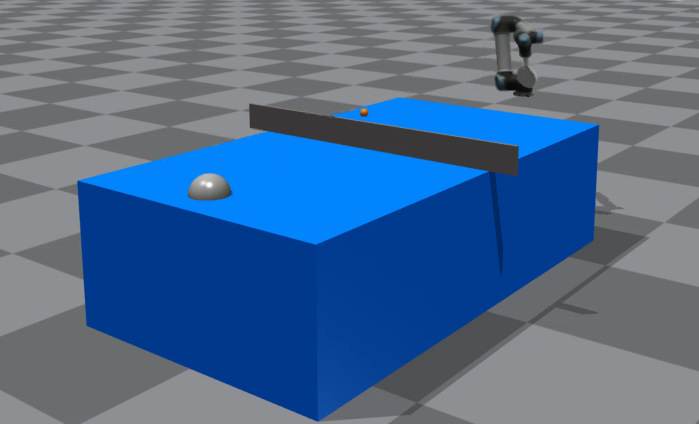}}
    \subfloat[amp-step3]{\includegraphics[width=0.14\textwidth]{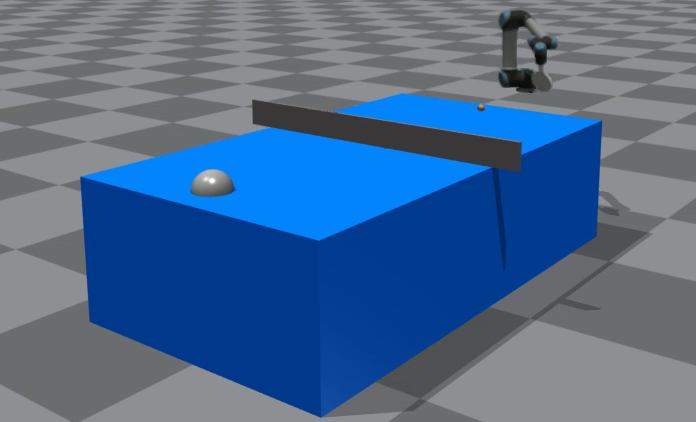}}
    \subfloat[amp-step4]{\includegraphics[width=0.14\textwidth]{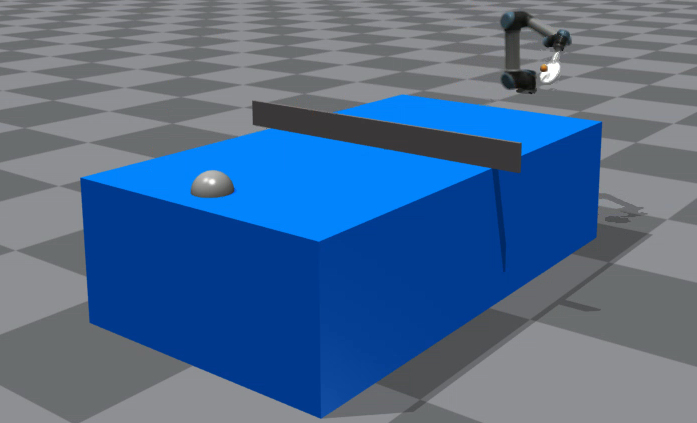}}
    \subfloat[amp-step5]{\includegraphics[width=0.14\textwidth]{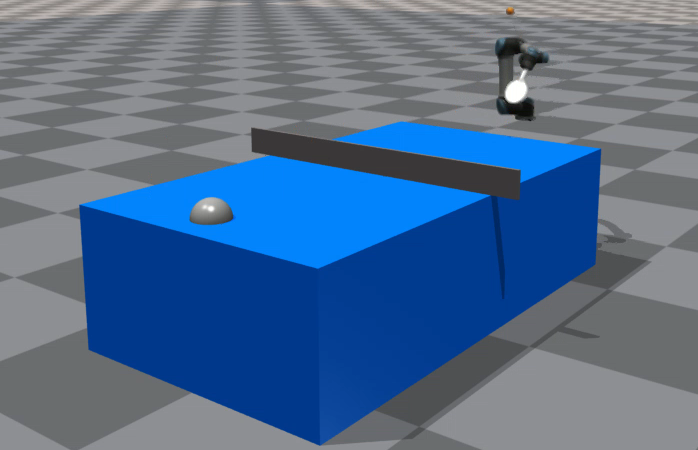}}
    \subfloat[amp-step6]{\includegraphics[width=0.14\textwidth]{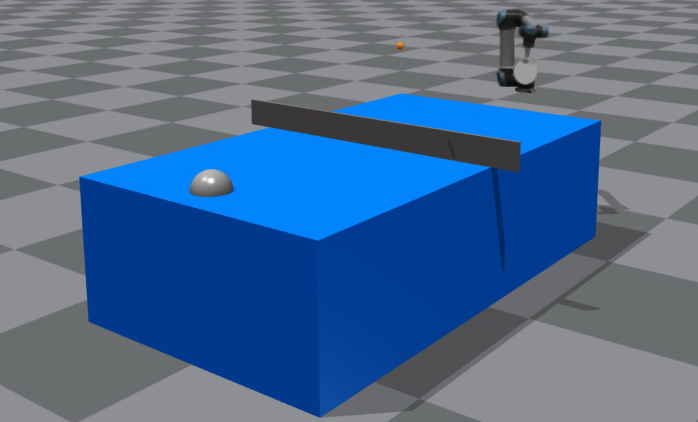}}
    \subfloat[amp-step7]{\includegraphics[width=0.14\textwidth]{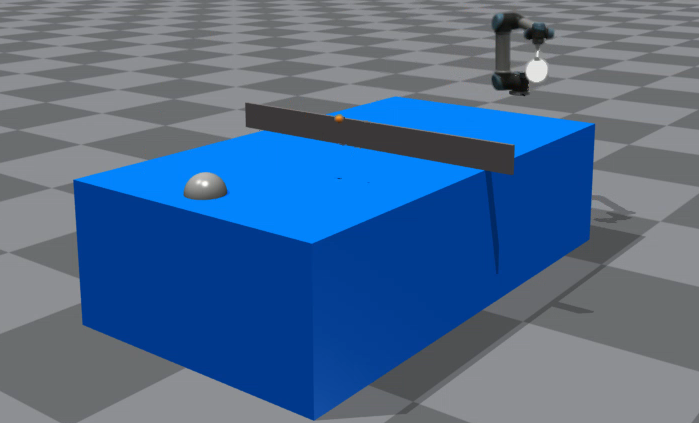}} \\
    \subfloat[motion-step1]{\includegraphics[width=0.14\textwidth]{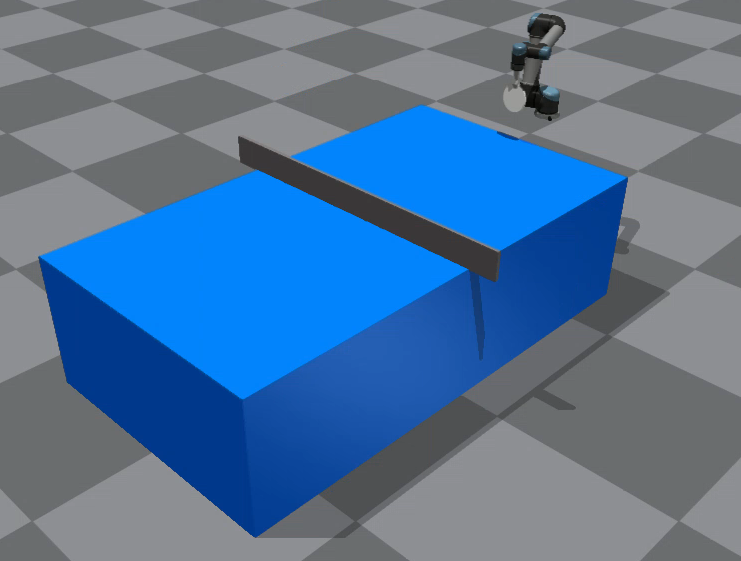}}
    \subfloat[motion-step2]{\includegraphics[width=0.14\textwidth]{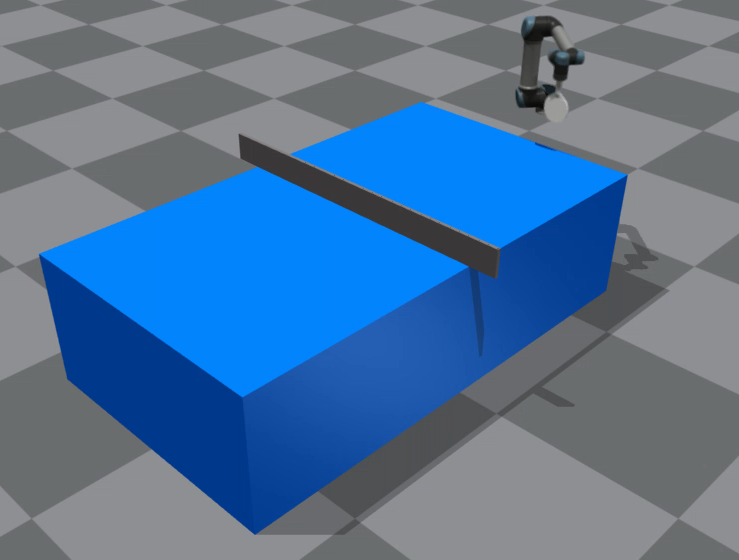}}
    \subfloat[motion-step3]{\includegraphics[width=0.14\textwidth]{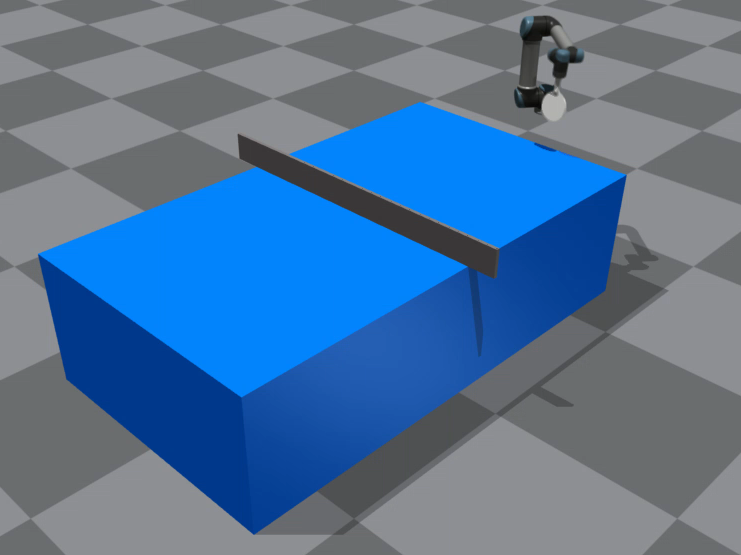}}
    \subfloat[motion-step4]{\includegraphics[width=0.14\textwidth]{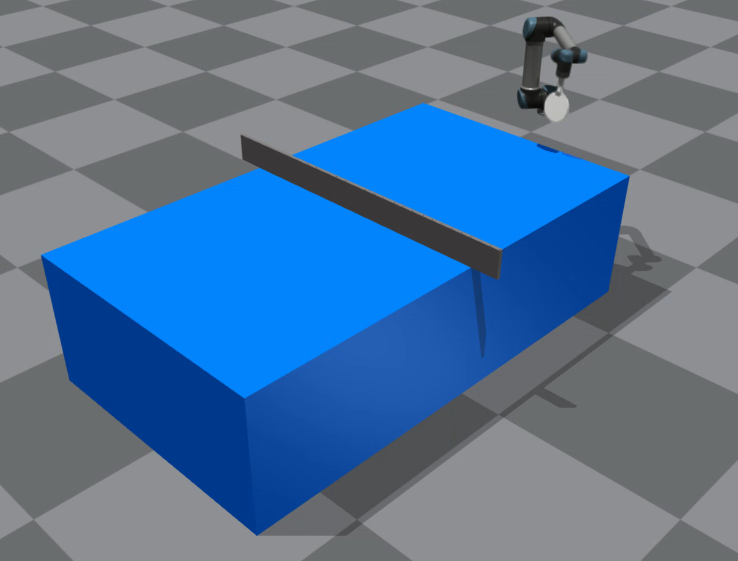}}
    \subfloat[motion-step5]{\includegraphics[width=0.14\textwidth]{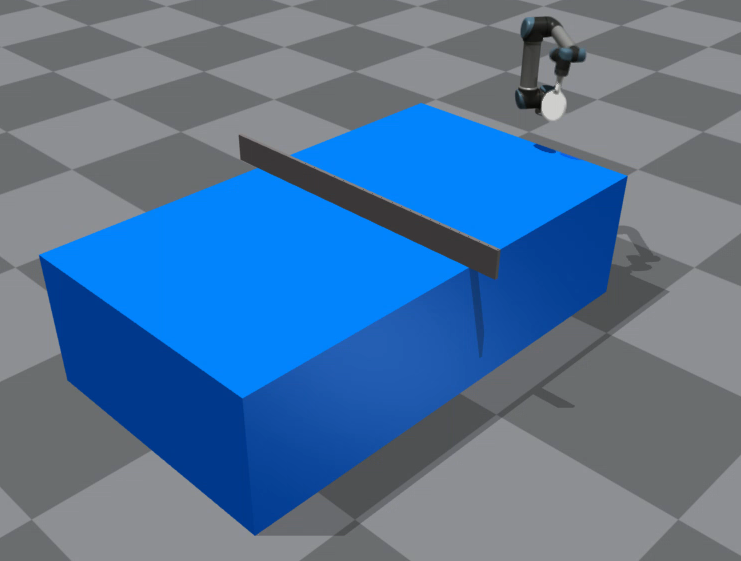}}
    \subfloat[motion-step6]{\includegraphics[width=0.14\textwidth]{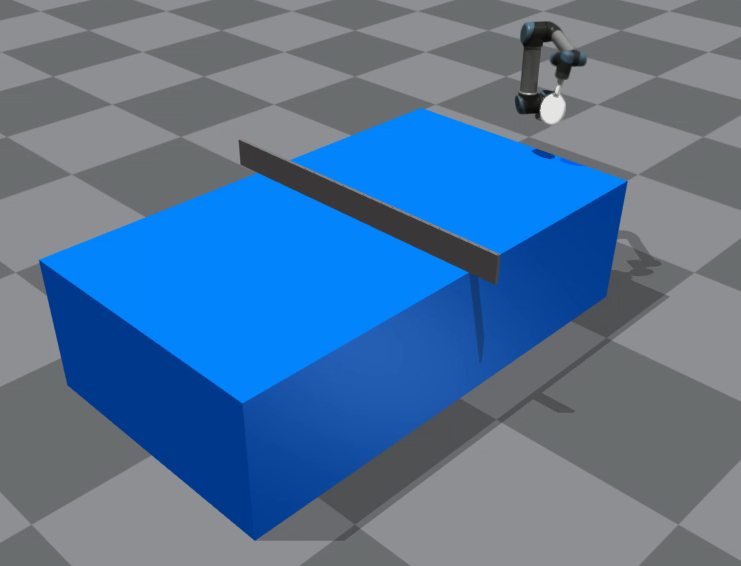}}
    \subfloat[motion-step7]{\includegraphics[width=0.14\textwidth]{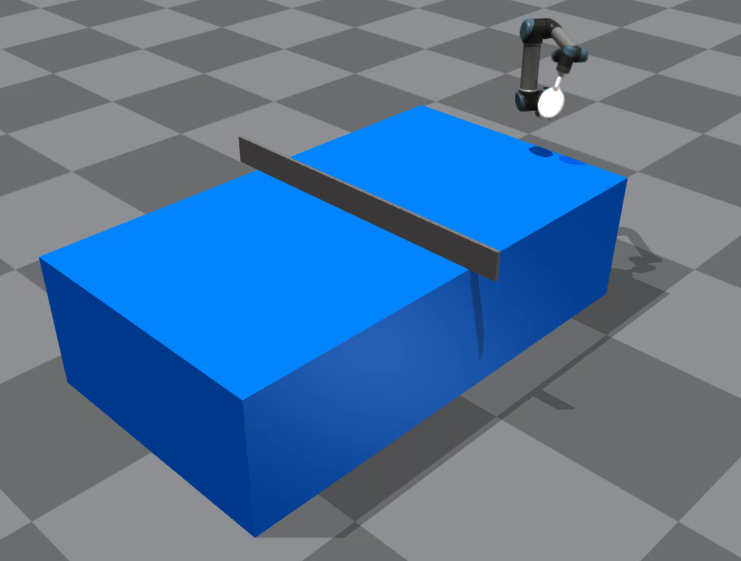}}
    \caption{Visualization of comparison of playing style of PPO and our method. We also display a motion clip to show the AMP has learned the style from reference motion.}
    \label{fig:vis-ppo-amp}
\end{figure*}

\section{Experiment}
We begin by detailing our experimental setup. Subsequently, we will: 
a) compare our method with the baseline PPO method both quantitatively and visually;
b) illustrate the enhancements achieved through reference motion data augmentation; and 
c) conduct a sim2sim experiment to show the successful transfer of the policy learned via our method to a distinct domain, demonstrating its potential for real-world deployment.

\subsection{Experiment Setup}
In our experiments, we choose the Universal Robots UR5 as our robotic arm Our simulation environment is constructed with IsaacGym \cite{makoviychuk2021isaac}. For the experiment's setup, a table tennis bat is affixed to the UR5 robot's end-effector. The simulation environment is reset under any of the following conditions: if the ball exits the bounds of the table, if there's an illegal bounce, if any part of the robotic arm other than the bat makes contact with the ball, or if the robotic arm collides with the table. We also define a goal region with a radius of $0.25$m. Upon every environment reset, this region is resampled across the entire half-length of the table. The primary objective of adding this goal region is to direct the robot to strike the ball across the broadest possible area on the table. 

The simulation operates at a frequency of 120Hz, whereas the policy operates at 60Hz. 4096 environments are simulated in parallel on a single Nvidia RTX3090 GPU, while the interactions between the ball, bat, and table are simulated on the CPU. For modeling the policy, value function, and discriminator, we utilize three separate multi-layer perceptrons (MLPs). The Exponential Linear Unit (ELU) activation function \cite{clevert2015fast} is employed. The hidden dimensions are [512, 256, 128] for both the policy and value function, while for the discriminator, they are [512, 256]. We employ the proximal policy optimization method \cite{schulman2017proximal} to train our policy. Over the course of training, it processes around 131 million samples. This translates to roughly 606 hours of simulated time, which corresponds to about 6 hours in real-world time.

\subsection{Playing Style Comparison}
\noindent\textbf{Visual Comparison}\quad We first trained two policies using PPO and our method with augmented reference motion dataset separately. Both policies were then tested in simulation with randomly incoming balls. Figure \ref{fig:vis-ppo-amp} illustrates the playing styles of the policy trained using PPO from scratch versus that trained with AMP. From the visualization, it is evident that the policy trained from scratch using PPO can display unrealistic behaviors. Even though this policy can manage to return the ball in the simulation, its implausible trajectory indicates that deploying this policy in real world is nearly impossible. Conversely, the policy trained with our method showcases more feasible and logical motions. And it also captured the style of our reference motion clip. Due to the smaller range of the action it displays, deploying this policy in real world is much more practical. We can also see that our trained policy has successfully captured the essence of input reference motions and is able to adjust the velocity to downstream tasks. 

\noindent\textbf{Quantitative Comparison}\quad To quantitatively assess these two policies, we evaluated their success rates in returning randomly spawned balls. We executed each policy in an identical environment using 10 distinct random seeds, amounting to \(200\) spawnings for each seed. The success rate was then averaged across these 10 seeds to obtain the mean success rate. The findings are presented in Table \ref{tab:result-comp}. From the results, it is evident that the policy trained with our method is not only robust but can also adeptly return balls from varied directions and velocities. With a success rate of \(93\%\), which closely parallels the \(97\%\) of the PPO policy, demonstrating that its performance is maintained quite well.

\begin{table}[ht]
\caption{Success rate comparison of policies trained with different methods.}
\begin{center}
\begin{tabular}{cccc}
\toprule
\textbf{Method} & \textbf{Total Attempts} & \textbf{Mean Success} & \textbf{Success Rate} \\
PPO & $200$ & $192.53$ & $96.27\%$ \\
AMP & $200$ & $163.22$ & $81.61\%$ \\
Our Method & $200$ & $180.64$ & $90.32\%$ \\
\bottomrule
\end{tabular}
\end{center}
\label{tab:result-comp}
\end{table}

To assess the similarity between states produced by our trained action and the reference motion, we adopt Dynamic Time Warping (DTW) as the metric \cite{muller2007dynamic}. We choose DTW due to the considerable variation in length and speed between the action generated by our trained policy and the reference motion, and DTW is primarily designed to measure the similarity between two sequences with potential differences in speed or length. In our experiment, we input the sequence of joint positions of our policy and reference motion respectively. Our reference motion comprises three main components: moving forward, leftward, and rightward. Accordingly, we introduce three types of balls moving in these directions and evaluate the states against the respective reference motions. The findings are presented in Table \ref{tab:dtw-comp}. The DTW distance, when compared to policies trained from scratch using PPO, is notably lower, demonstrating the effectiveness of our approach.

\begin{table}[ht]
\caption{Comparison of DTW distance for PPO and our method. Lower is better.}
\begin{center}
\begin{tabular}{cccc}
\toprule
\textbf{\quad} & \textbf{Forward} & \textbf{Leftward} & \textbf{Rightward} \\
\textbf{PPO} & $148.23$ & $263.77$ & $403.85$ \\
\textbf{Our Method} & \textbf{62.73} & \textbf{133.22} & \textbf{243.56} \\
\bottomrule
\end{tabular}
\end{center}
\label{tab:dtw-comp}
\end{table}

Both visual and quantitative analyses clearly indicate that our method enables the robot to exhibit realistic and plausible behaviors, while performance levels remain largely consistent.

\subsection{Data Augmentation}
We evaluated the efficacy of our motion data augmentation technique using success rate metrics. The overall results are provided in Table \ref{tab:result-comp}. A noticeable improvement is discernible after augmentation. The primary motivation behind our data augmentation is to condition our trained agent to adeptly handle balls traveling at high speeds. We subsequently measured the success rate of two agents when presented with low-speed balls ($4.0\sim4.8~m/s$) and high-speed balls (spanning from $5.8\sim6.6~m/s$). These findings are presented in Table \ref{tab:aug-comp}. 

For balls traveling at a reduced velocity, both the original AMP and our augmented method manage to return the ball with commendable efficiency, producing a success rate that surpasses \(90\%\). In contrast, when faced with balls spawned with augmented speeds, the agent trained via the original AMP exhibits a significant decline in its success rate, which is approximately \(70\%\). Our method, conversely, showcases minimal variation in performance, whose success rate is around $86\%$, differs little compared with low speed balls. The supplementary video further demonstrates the resilience and adaptability of our method to balls of varying velocities.

\begin{table}[ht]
\caption{Success rate comparison of policies trained with origianl AMP and our method on different ball speeds. (\textit{If the agent can properly return the spawned ball to the opposing side of the table, then it is considered a success, regardless whether the ball is landed in the goal region.})}
\begin{center}
\begin{tabular}{ccccc}
\toprule
\textbf{Method} & \textbf{Speed} & \textbf{Total Attempts} & \textbf{Mean} & \textbf{Success Rate} \\
AMP & Low & $200$ & $184.29$ & $92.15\%$ \\
Our Method & Low & $200$ & $183.89$ & $91.95\%$ \\
AMP & High & $200$ & $138.00$ & $69.00\%$ \\
Our method & High & $200$ & $171.00$ & $85.91\%$ \\
\bottomrule
\end{tabular}
\end{center}
\label{tab:aug-comp}
\end{table}

\subsection{Sim2Sim Transfer}
\begin{figure}[ht]
    \centering
    \subfloat[IsaacGym]{\includegraphics[width=0.24\textwidth]{imgs/ppo-final/frame1-edited.png}} 
    \subfloat[PyBullet]{\includegraphics[width=0.24\textwidth]{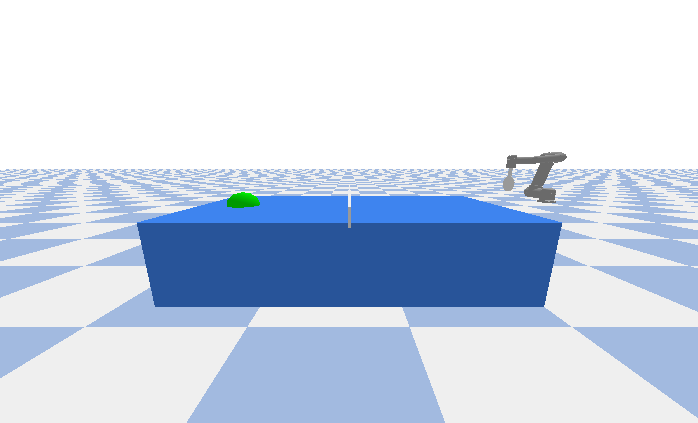}}
    \caption{IsaacGym environment and PyBullet environment.}
    \label{fig:vis-isaac-bullet}
\end{figure}
To demonstrate the potential of our method for real-world applications, we effectively transitioned it to an environment constructed on PyBullet \cite{coumans2019}. A visualization of two environments is shown in Figure \ref{fig:vis-isaac-bullet}. Specifically, we maintained uniformity in the key dynamics parameters, mirroring those used in IsaacGym. However, the difference between the simulation engines and the low level controllers and their different simulation methodologies leads to a gap in the displayed dynamics. Actually, this difference effectively mirrors the main discrepancy between simulations and real-world dynamics. With previously detailed domain randomization, we trained a policy with our method and subsequently tested this policy within the PyBullet framework. The success rates are shown in Table \ref{tab:sim2sim-comp}. We can see that after transferring, there is a huge decrease in performance compared with the our policy in the original domain. However, it still remains a fairly good performance with a success rate about $40\%$.

\begin{table}[ht]
\caption{Success rate comparison of policies tested in IsaacGym and PyBullet. (\textit{If the agent can properly return the spawned ball to the opposing side of the table, then it is considered a success, regardless whether the ball is landed in the goal region.})}
\begin{center}
\begin{tabular}{cccc}
\toprule
\textbf{Simulator} & \textbf{Total Attempts} & \textbf{Mean Success} & \textbf{Success Rate} \\
IsaacGym & $200$ & $142.78$ & $71.39\%$ \\
PyBullet & $200$ & $78.64$ & $39.32\%$ \\
\bottomrule
\end{tabular}
\end{center}
\label{tab:sim2sim-comp}
\end{table}

\section{Conclusions and Limitations}
In this study, we constructed a stylized table tennis agent using partial demonstrations derived from dragging-and-teaching methods. This agent produces actions that are more realistic compared to those generated by agents trained from scratch with PPO. We further incorporated data augmentation on the reference motion data, enhancing the agent's adaptability to balls of varying speeds. Additionally, we successfully transferred this agent to a domain with distinct dynamics, implemented using PyBullet.

Though we have effectively transferred the policy to a different domain, the evaluation of our policy on a real robot remains unsolved. Time constraints and our selection of high-speed cameras have precluded its transfer to an actual robot, which we consider as our subsequent step. Additionally, our current demonstrations, sourced from human-assisted dragging, are somewhat restricted compared to the diverse demonstrations available from online videos. Thus, there is potential in developing methodologies that enable our robot to learn styles from online table tennis match videos.

\bibliographystyle{IEEEtran}
\bibliography{IEEEabrv,ref}

\begin{thebibliography}{10}
\providecommand{\url}[1]{#1}
\csname url@rmstyle\endcsname
\providecommand{\newblock}{\relax}
\providecommand{\bibinfo}[2]{#2}
\providecommand\BIBentrySTDinterwordspacing{\spaceskip=0pt\relax}
\providecommand\BIBentryALTinterwordstretchfactor{4}
\providecommand\BIBentryALTinterwordspacing{\spaceskip=\fontdimen2\font plus
\BIBentryALTinterwordstretchfactor\fontdimen3\font minus \fontdimen4\font\relax}
\providecommand\BIBforeignlanguage[2]{{%
\expandafter\ifx\csname l@#1\endcsname\relax
\typeout{** WARNING: IEEEtran.bst: No hyphenation pattern has been}%
\typeout{** loaded for the language `#1'. Using the pattern for}%
\typeout{** the default language instead.}%
\else
\language=\csname l@#1\endcsname
\fi
#2}}

\bibitem{gao2020robotic}
W.~Gao, L.~Graesser, K.~Choromanski, X.~Song, N.~Lazic, P.~Sanketi, V.~Sindhwani, and N.~Jaitly, ``Robotic table tennis with model-free reinforcement learning,'' in \emph{2020 IEEE/RSJ International Conference on Intelligent Robots and Systems (IROS)}.\hskip 1em plus 0.5em minus 0.4em\relax IEEE, 2020, pp. 5556--5563.

\bibitem{mahjourian2019hierarchical}
R.~Mahjourian, R.~Miikkulainen, N.~Lazic, S.~Levine, and N.~Jaitly, ``Hierarchical policy design for sample-efficient learning of robot table tennis through self-play,'' 2019.

\bibitem{dambrosio2023robotic}
D.~B. D'Ambrosio, J.~Abelian, S.~Abeyruwan, M.~Ahn, A.~Bewley, J.~Boyd, K.~Choromanski, O.~Cortes, E.~Coumans, T.~Ding, W.~Gao, L.~Graesser, A.~Iscen, N.~Jaitly, D.~Jain, J.~Kangaspunta, S.~Kataoka, G.~Kouretas, Y.~Kuang, N.~Lazic, C.~Lynch, R.~Mahjourian, S.~Q. Moore, T.~Nguyen, K.~Oslund, B.~J. Reed, K.~Reymann, P.~R. Sanketi, A.~Shankar, P.~Sermanet, V.~Sindhwani, A.~Singh, V.~Vanhoucke, G.~Vesom, and P.~Xu, ``Robotic table tennis: A case study into a high speed learning system,'' in \emph{Robotics: Science and Systems}, 2023.

\bibitem{monte2018zhu}
Y.~Zhu, Y.~Zhao, L.~Jin, J.~Wu, and R.~Xiong, ``Towards high level skill learning: Learn to return table tennis ball using monte-carlo based policy gradient method,'' in \emph{2018 IEEE International Conference on Real-time Computing and Robotics (RCAR)}, 2018, pp. 34--41.

\bibitem{optstrike2015huang}
Y.~Huang, B.~Schölkopf, and J.~Peters, ``Learning optimal striking points for a ping-pong playing robot,'' in \emph{2015 IEEE/RSJ International Conference on Intelligent Robots and Systems (IROS)}, 2015, pp. 4587--4592.

\bibitem{balance2011sun}
Y.~Sun, R.~Xiong, Q.~Zhu, J.~Wu, and J.~Chu, ``Balance motion generation for a humanoid robot playing table tennis,'' in \emph{2011 11th IEEE-RAS International Conference on Humanoid Robots}, 2011, pp. 19--25.

\bibitem{gao2022optimal}
Y.~Gao, J.~Tebbe, and A.~Zell, ``Optimal stroke learning with policy gradient approach for robotic table tennis,'' \emph{Applied Intelligence}, pp. 1--14, 2022.

\bibitem{gao2022model}
------, ``A model-free approach to stroke learning for robotic table tennis,'' in \emph{2022 International Joint Conference on Neural Networks (IJCNN)}.\hskip 1em plus 0.5em minus 0.4em\relax IEEE, 2022, pp. 1--8.

\bibitem{tebbe2021sample}
J.~Tebbe, L.~Krauch, Y.~Gao, and A.~Zell, ``Sample-efficient reinforcement learning in robotic table tennis,'' in \emph{2021 IEEE international conference on robotics and automation (ICRA)}.\hskip 1em plus 0.5em minus 0.4em\relax IEEE, 2021, pp. 4171--4178.

\bibitem{yang2021ball}
L.~Yang, H.~Zhang, X.~Zhu, and X.~Sheng, ``Ball motion control in the table tennis robot system using time-series deep reinforcement learning,'' \emph{IEEE Access}, vol.~9, pp. 99\,816--99\,827, 2021.

\bibitem{buchler2022learning}
D.~B{\"u}chler, S.~Guist, R.~Calandra, V.~Berenz, B.~Sch{\"o}lkopf, and J.~Peters, ``Learning to play table tennis from scratch using muscular robots,'' \emph{IEEE Transactions on Robotics}, vol.~38, no.~6, pp. 3850--3860, 2022.

\bibitem{ding2022goalseye}
T.~Ding, L.~Graesser, S.~W. Abeyruwan, D.~B. D'Ambrosio, A.~Shankar, P.~Sermanet, P.~R. Sanketi, and C.~H. Lynch, ``Goalseye: Learning high speed precision table tennis on a physical robot,'' 2022.

\bibitem{abeyruwan2023sim2real}
S.~W. Abeyruwan, L.~Graesser, D.~B. D’Ambrosio, A.~Singh, A.~Shankar, A.~Bewley, D.~Jain, K.~M. Choromanski, and P.~R. Sanketi, ``i-sim2real: Reinforcement learning of robotic policies in tight human-robot interaction loops,'' in \emph{Conference on Robot Learning}.\hskip 1em plus 0.5em minus 0.4em\relax PMLR, 2023, pp. 212--224.

\bibitem{pathak2019self}
D.~Pathak, D.~Gandhi, and A.~Gupta, ``Self-supervised exploration via disagreement,'' in \emph{International conference on machine learning}.\hskip 1em plus 0.5em minus 0.4em\relax PMLR, 2019, pp. 5062--5071.

\bibitem{billard2008survey}
A.~Billard, S.~Calinon, R.~Dillmann, and S.~Schaal, ``Survey: Robot programming by demonstration,'' Springrer, Tech. Rep., 2008.

\bibitem{schaal1996learning}
S.~Schaal, ``Learning from demonstration,'' \emph{Advances in neural information processing systems}, vol.~9, 1996.

\bibitem{goodfellow2014generative}
I.~Goodfellow, J.~Pouget-Abadie, M.~Mirza, B.~Xu, D.~Warde-Farley, S.~Ozair, A.~Courville, and Y.~Bengio, ``Generative adversarial nets,'' \emph{Advances in neural information processing systems}, vol.~27, 2014.

\bibitem{ho2016generative}
J.~Ho and S.~Ermon, ``Generative adversarial imitation learning,'' \emph{Advances in neural information processing systems}, vol.~29, 2016.

\bibitem{liu2010sampling}
L.~Liu, K.~Yin, M.~Van~de Panne, T.~Shao, and W.~Xu, ``Sampling-based contact-rich motion control,'' in \emph{ACM SIGGRAPH 2010 papers}, 2010, pp. 1--10.

\bibitem{liu2016guided}
L.~Liu, M.~V.~D. Panne, and K.~Yin, ``Guided learning of control graphs for physics-based characters,'' \emph{ACM Transactions on Graphics (TOG)}, vol.~35, no.~3, pp. 1--14, 2016.

\bibitem{peng2018deepmimic}
X.~B. Peng, P.~Abbeel, S.~Levine, and M.~Van~de Panne, ``Deepmimic: Example-guided deep reinforcement learning of physics-based character skills,'' \emph{ACM Transactions On Graphics (TOG)}, vol.~37, no.~4, pp. 1--14, 2018.

\bibitem{fu2017learning}
J.~Fu, K.~Luo, and S.~Levine, ``Learning robust rewards with adversarial inverse reinforcement learning,'' \emph{arXiv preprint arXiv:1710.11248}, 2017.

\bibitem{peng2021amp}
X.~B. Peng, Z.~Ma, P.~Abbeel, S.~Levine, and A.~Kanazawa, ``Amp: Adversarial motion priors for stylized physics-based character control,'' \emph{ACM Transactions on Graphics (TOG)}, vol.~40, no.~4, pp. 1--20, 2021.

\bibitem{escontrela2022adversarial}
A.~Escontrela, X.~B. Peng, W.~Yu, T.~Zhang, A.~Iscen, K.~Goldberg, and P.~Abbeel, ``Adversarial motion priors make good substitutes for complex reward functions,'' in \emph{2022 IEEE/RSJ International Conference on Intelligent Robots and Systems (IROS)}.\hskip 1em plus 0.5em minus 0.4em\relax IEEE, 2022, pp. 25--32.

\bibitem{wang2023amp}
Y.~Wang, Z.~Jiang, and J.~Chen, ``Amp in the wild: Learning robust, agile, natural legged locomotion skills,'' \emph{arXiv preprint arXiv:2304.10888}, 2023.

\bibitem{li2023learning}
C.~Li, M.~Vlastelica, S.~Blaes, J.~Frey, F.~Grimminger, and G.~Martius, ``Learning agile skills via adversarial imitation of rough partial demonstrations,'' in \emph{Conference on Robot Learning}.\hskip 1em plus 0.5em minus 0.4em\relax PMLR, 2023, pp. 342--352.

\bibitem{scherzinger2017forward}
S.~Scherzinger, A.~Roennau, and R.~Dillmann, ``Forward dynamics compliance control (fdcc): A new approach to cartesian compliance for robotic manipulators,'' in \emph{2017 IEEE/RSJ International Conference on Intelligent Robots and Systems (IROS)}.\hskip 1em plus 0.5em minus 0.4em\relax IEEE, 2017, pp. 4568--4575.

\bibitem{mao2017least}
X.~Mao, Q.~Li, H.~Xie, R.~Y. Lau, Z.~Wang, and S.~Paul~Smolley, ``Least squares generative adversarial networks,'' in \emph{Proceedings of the IEEE international conference on computer vision}, 2017, pp. 2794--2802.

\bibitem{gulrajani2017improved}
I.~Gulrajani, F.~Ahmed, M.~Arjovsky, V.~Dumoulin, and A.~C. Courville, ``Improved training of wasserstein gans,'' \emph{Advances in neural information processing systems}, vol.~30, 2017.

\bibitem{pinto2017asymmetric}
L.~Pinto, M.~Andrychowicz, P.~Welinder, W.~Zaremba, and P.~Abbeel, ``Asymmetric actor critic for image-based robot learning,'' \emph{arXiv preprint arXiv:1710.06542}, 2017.

\bibitem{makoviychuk2021isaac}
V.~Makoviychuk, L.~Wawrzyniak, Y.~Guo, M.~Lu, K.~Storey, M.~Macklin, D.~Hoeller, N.~Rudin, A.~Allshire, A.~Handa, \emph{et~al.}, ``Isaac gym: High performance gpu-based physics simulation for robot learning,'' \emph{arXiv preprint arXiv:2108.10470}, 2021.

\bibitem{clevert2015fast}
D.-A. Clevert, T.~Unterthiner, and S.~Hochreiter, ``Fast and accurate deep network learning by exponential linear units (elus),'' \emph{arXiv preprint arXiv:1511.07289}, 2015.

\bibitem{schulman2017proximal}
J.~Schulman, F.~Wolski, P.~Dhariwal, A.~Radford, and O.~Klimov, ``Proximal policy optimization algorithms,'' \emph{arXiv preprint arXiv:1707.06347}, 2017.

\bibitem{muller2007dynamic}
M.~M{\"u}ller, ``Dynamic time warping,'' \emph{Information retrieval for music and motion}, pp. 69--84, 2007.

\bibitem{coumans2019}
E.~Coumans and Y.~Bai, ``Pybullet, a python module for physics simulation for games, robotics and machine learning,'' \url{http://pybullet.org}, 2016--2019.

\end{thebibliography}

\end{document}